\documentclass[letterpaper]{article} 
\usepackage{aaai24}  
\usepackage{times}  
\usepackage{helvet}  
\usepackage{courier}  
\usepackage[hyphens]{url}  
\usepackage{graphicx} 
\usepackage{natbib}  
\usepackage{caption} 
\usepackage{algorithm}
\usepackage{algorithmic}

\usepackage{newfloat}
\usepackage{listings}
\DeclareCaptionStyle{ruled}{labelfont=normalfont,labelsep=colon,strut=off} 
\floatstyle{ruled}
\newfloat{listing}{tb}{lst}{}
\floatname{listing}{Listing}
\usepackage{color}
\usepackage{booktabs}
\usepackage{amsmath}
\usepackage{amsfonts,amssymb}

\title{Instance-aware Multi-Camera 3D Object Detection with Structural Priors \\ Mining and Self-Boosting Learning}
\author{
    Yang Jiao\textsuperscript{\rm 1,2}, Zequn Jie\textsuperscript{\rm 3}, Shaoxiang Chen\textsuperscript{\rm 3}, Lechao Cheng\textsuperscript{\rm 4}, Jingjing Chen\textsuperscript{\rm 1,2}$^{\dagger}$,\\ Lin Ma\textsuperscript{\rm 3}$^{\dagger}$\thanks{$\dagger$Corresponding authors}, Yu-Gang Jiang\textsuperscript{\rm 1,2}
}
\affiliations{
    \textsuperscript{\rm 1}Shanghai Key Lab of Intell. Info. Processing, School of CS, Fudan University\\
    \textsuperscript{\rm 2}Shanghai Collaborative Innovation Center on Intelligent Visual Computing\\
    \textsuperscript{\rm 3}Meituan \textsuperscript{\rm 4}Zhejiang Lab

}

\usepackage{bibentry}

\makeatletter
\def\thanks#1{\protected@xdef\@thanks{\@thanks
        \protect\footnotetext{#1}}}
\makeatother

\begin{document}

\maketitle

\begin{abstract}
Camera-based bird-eye-view (BEV) perception paradigm has made significant progress in the autonomous driving field. 
Under such a paradigm, accurate BEV representation construction relies on reliable depth estimation for multi-camera images. However, existing approaches exhaustively predict depths for every pixel without prioritizing objects, which are precisely the entities requiring detection in the 3D space. To this end, we propose IA-BEV, which integrates image-plane instance awareness into the depth estimation process within a BEV-based detector. 
First, a category-specific structural priors mining approach is proposed for enhancing the efficacy of monocular depth generation. Besides, a self-boosting learning strategy is further proposed to encourage the model to place more emphasis on challenging objects in computation-expensive temporal stereo matching. Together they provide advanced depth estimation results for high-quality BEV features construction, benefiting the ultimate 3D detection. The proposed method achieves state-of-the-art performances on the challenging nuScenes benchmark, and extensive experimental results demonstrate the effectiveness of our designs.
\end{abstract}

\section{Introduction}

In recent years, there has been a surge of research interest in multi-camera 3D object detection within the autonomous driving field~\cite{huang2021bevdet,li2022bevformer,li2023bevdepth,feng2023aedet}. Compared with LiDAR, the camera excels at capturing object semantics and enjoys the advantage of a lower deployment cost. The recent trend in this field is to transform the multi-view image features to a unified Bird's-Eye-View (BEV) space for the subsequent perception. This paradigm facilitates aligning signals from multiple sensors and timestamps in the BEV space, serving as a generic representation for downstream tasks such as detection~\cite{jiao2023msmdfusion,li2023bevnext}, map segmentation~\cite{xie2022m} and motion planning~\cite{hu2023planning}.

Within the BEV-based perception pipeline, depth estimation plays a pivotal role in the perspective projection from the image view to BEV. Pioneering methods estimate depth from monocular images either implicitly~\cite{li2022bevformer} or explicitly~\cite{li2023bevdepth}. Motivated by the success of the multi-view stereo technique~\cite{yao2018mvsnet,wei2021aa}, follow-up approaches~\cite{li2022bevstereo,park2022time} leverage consecutive camera frames to construct cost volume for stereo matching. Benefiting from enhanced depth estimation, these methods enjoy high-quality BEV features and thus achieve remarkable detection performances.

\begin{figure}[t]
\centering
\includegraphics[width=\columnwidth]{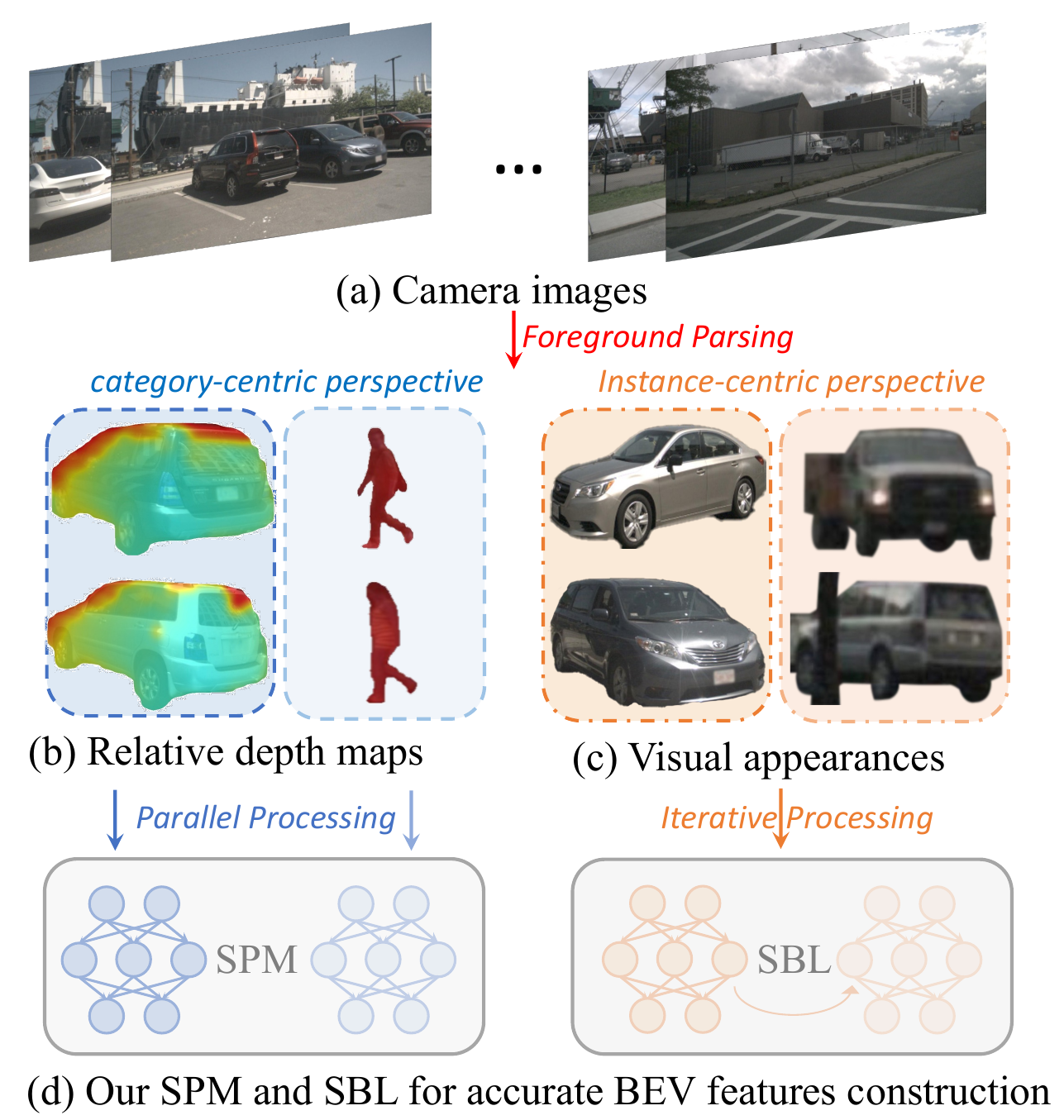} 
\caption{The overall workflow of the proposed IA-BEV. 
In (b), we first use gaussian kernel to densely fill depth values for each pixel, then calculate relative depths by dividing the maximum depth of the corresponding objects. All objects illustrated in (b) and (c) are extracted from nuScenes images and resized to the same scale for clarity.}
\label{fig1}
\end{figure}

Despite significant advancements, existing methods~\cite{li2023bevdepth,li2022bevstereo} treat every pixel equally, neglecting the inherent properties encapsulated in foreground objects. 
In fact, foreground objects can exhibit consistency within a category and heterogeneity among instances, which we find can be utilized to improve depth estimation.
On the one hand, objects of the same semantic category share similar structural priors, which manifest in two key aspects. 
1) Object scales shown in the image are correlated with their real depths, and this correlation is usually coherent for objects of the same semantic category and varies across categories. For example, the scales of cars in image are inversely proportional to their real depths, but even with the same depth, a car and a pedestrian can exhibit significantly different scales. 
2) Objects of the same semantic category have consistent inner-geometric structures. As shown in Fig~\ref{fig1}(b), when observed in isolation from the image plane, the objects of the same category (car) have similar distributions of relative depths.
On the other hand, for individual objects, their visual appearance can vary dramatically even within the same category due to different resolutions and occlusion statuses. Consequently, the complexity of depth estimation for different object instances also varies. 
As demonstrated in Fig~\ref{fig1}(c), cars on the left column contain more precise textures and shape details versus those on the right column, thus reducing the ambiguity of the challenging depth estimation. Although some approaches~\cite{chu2023oa,wang2023object} have explored 2D object priors for 3D object detection, they primarily leverage detected 2D objects after the perspective projection, thereby ignoring their potential to improve depth estimation for enhanced BEV feature construction.

Motivated by the aforementioned observations and to overcome the limitation of existing methods, we propose IA-BEV, which exploits 2D instance awareness to enhance the depth estimation process in the BEV-based detector. As shown in Fig~\ref{fig1}(d), our IA-BEV initially parses a scene into individual objects and then leverages their intrinsic properties to assist both monocular and stereo depth estimation with novelly devised Structural Priors Mining approach (SPM) and Self-Boosting Learning strategy (SBL), respectively. Within SPM, objects belonging to the same or similar semantic categories are grouped and processed by respective lightweight depth decoders to better exploit structural priors. However, expecting these parallel decoders to actively learn category-specific patterns with only grouped input poses significant optimization challenges, resulting in suboptimal performance. To address this, we explicitly encode object scale properties as additional inputs and apply two instance-aware loss functions to supervise the rough instance absolute depth and the fine-grained inner-object relative depth predictions.
In contrast to SPM, SBL operates in a class-agnostic manner, which focuses on iteratively distinguishing and emphasizing challenging objects.
Within each iteration, objects are first partitioned into two groups according to their stereo-matching uncertainty. Subsequently, the group with higher uncertainty, indicating inaccurate estimation, is further boosted in the subsequent iteration. Thanks to the gradually sparser foreground regions addressed in the later iterations, we can set denser depth hypotheses within the realm of uncertainty for more comprehensive stereo matching on the selected challenging samples.
Finally, on the basis of the combined depth estimates from both SPM and SBL, the conventional view transformation process is conducted to construct BEV features for the ultimate detection. 

In summary, our contributions are three-fold: (1) We propose IA-BEV, which enhances the depth estimation process within the BEV perception pipeline via exploiting 2D instance awareness. (2) Within IA-BEV, a Structural Priors Mining approach (SPM) and a Self-Boosting Learning strategy (SBL) are introduced to exploit object intrinsic properties to promote monocular and stereo depth estimation, respectively. (3) Our IA-BEV achieves significant improvements over the strong BEVDepth baseline and state-of-the-art performances among all methods that also utilize two keyframes on the challenging nuScenes benchmark. 

\section{Related Work}
\subsection{Depth Estimation}
Estimating depths from camera images has been a classical topic in computer vision. Contemporary research can be grouped into monocular
and stereo depth estimation approaches. Monocular depth estimation aims to predict depths from a single image. Mainstream methods~\cite{bhat2021adabins,poggi2020uncertainty,li2022depthformer} in this line adopt an encoder-decoder pipeline to directly predict the depth values or distribution at the input resolution. However, monocular depth estimation is a longstanding ill-posed problem due to its inherent scale ambiguity. As an alternative, stereo depth estimation is based on multi-view image inputs to construct the cost volume to learn the pixel-to-pixel matching behaviors to meet the epipolar geometry constraints~\cite{yao2018mvsnet,peng2022rethinking,shen2021cfnet}. As a key step to link the 2D and 3D space, depth estimation techniques have been extensively adopted in modern autonomous-driving field 3D detectors~\cite{sun2020disp,li2023bevdepth,li2022bevstereo}. However, due to the scene layout complexity and supervision signals sparsity in the outdoor scenarios, the depth estimation quality in these detectors remains unsatisfactory.

\begin{figure*}[t]
\centering
\includegraphics[width=\textwidth]{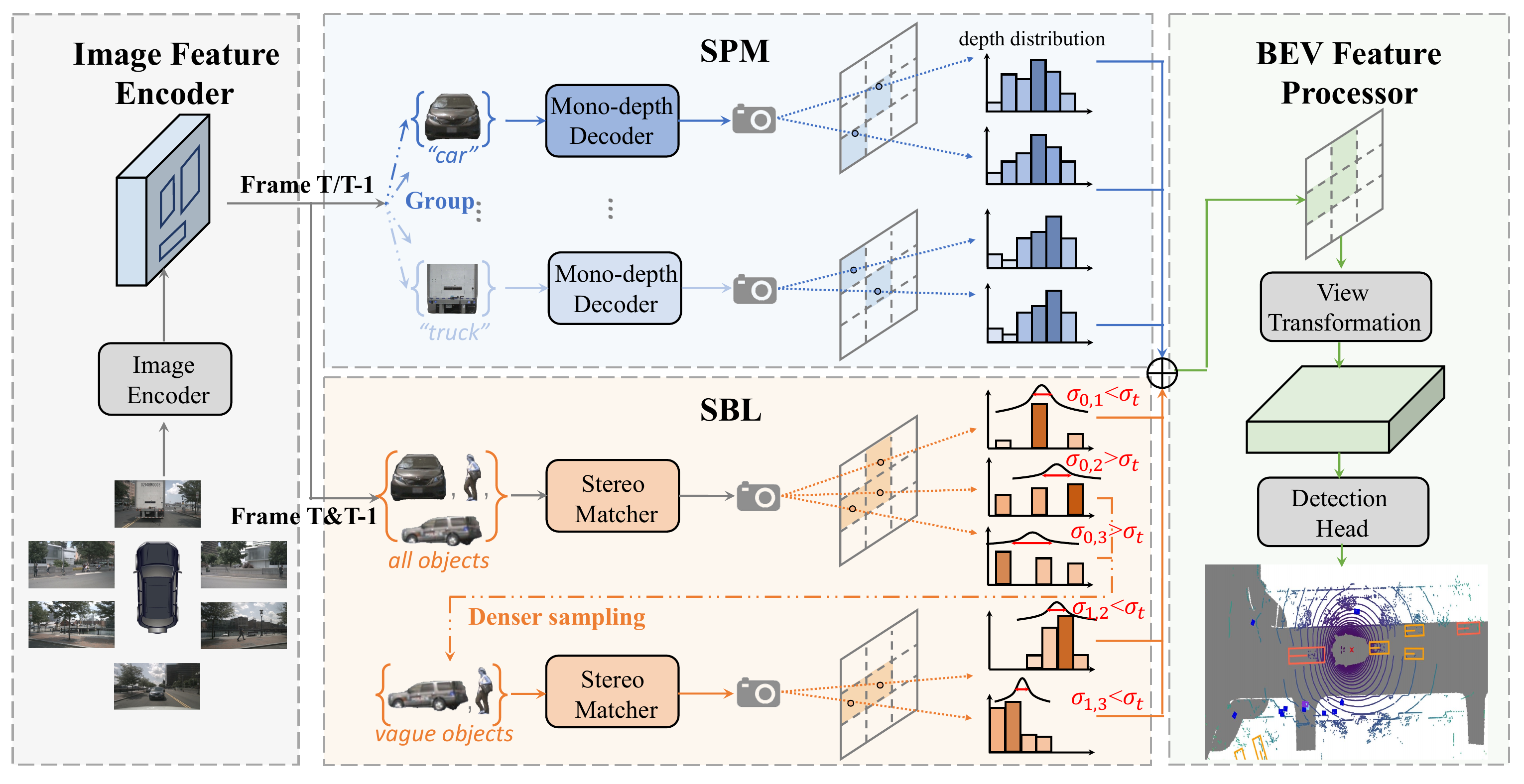} 
\caption{The detailed designs of the proposed IA-BEV. Given images collected from multi-view cameras, foreground objects are first parsed by off-the-shelf 2D scene parsers. Then, these objects, together with image features, are fed into the proposed SPM and SBL for effective depth estimation by exploring object properties from category-centric and instance-centric perspectives, respectively. Finally, the outputs from both SPM and SBL are merged, resulting in the ultimate image depths, which will be used for the conventional view transformation and BEV-based detection. Frame $T$ and $T-1$ are individually fed into SPM, while they are supplied to SBL simultaneously as the stereo matching here requires temporal multi-view information.}
\label{fig2}
\end{figure*}

\subsection{Multi-camera 3D Object Detection}
Modern BEV-based 3D detectors can be categorized into two paradigms. The first paradigm transforms image features to the BEV space via the Lift-Splat-Shoot~\cite{philion2020lift} technique. The pioneering work, BEVDet~\cite{huang2021bevdet}, first implements the complete LSS-based detection pipeline. Follow-up approaches~\cite{li2023bevdepth,li2022bevstereo,park2022time} enhance the depth estimation by introducing depth supervision or leveraging multi-frame information. Another line of work projects 3D object queries to multi-view image planes to collect useful image features. DETR3D~\cite{wang2022detr3d} and PETR~\cite{liu2022petr} extend DETR~\cite{carion2020end} detector to mutli-camera 3D object detection. BEVFormer-series methods~\cite{li2022bevformer,yang2023bevformer} further incorporate both spatial and temporal cues for more robust detection. On the basis of these two fundamental paradigms, some recent studies have explored 2D object priors to assist 3D detection. OA-BEV~\cite{chu2023oa} lifts detected regions as foreground objects from the image plane to 3D space, and processes them with a voxel encoder to generate object-aware BEV features as the augmentation of the original ones. MV2D~\cite{wang2023object} utilizes 2D object features and detections to directly predict 3D detection results. Instead of relying on 2D object detections for late fusion or final decision, our IA-BEV aims to harness object inherent properties in early depth estimation, enabling the construction of high-quality BEV features.

\section{Method}
As shown in Fig~\ref{fig2}, the proposed IA-BEV comprises four key components: a feature encoder responsible for extracting image features and parsing foreground objects, the proposed Structural Priors Mining approach (SPM) which enhances the monocular depth estimation by leveraging structural consistency of the same category of objects, the proposed Self-Boosting Learning strategy (SBL) which emphasizes vague objects during stereo depth estimation, and finally, a BEV feature encoder utilized for rendering features and detecting objects in the BEV space. In the following sections, we will elaborate on each of them.

\subsection{Feature Encoder}
Given images collected from multi-view cameras, we extract image features with a prevalent backbone like ResNet-50~\cite{he2016deep} or ConvNeXt~\cite{liu2022convnet}. Meanwhile, we use the off-the-shelf instance segmentor~\cite{zhou2021probabilistic} to parse foreground objects $\mathcal{O}=\{(F_i, b_i)\}_{i=1}^{N}$, where $F_i$ includes features of all pixels belonging to the current object, $b_i$ refers to box parameters, and $N$ is the number of segemented foreground objects. Note that here we keep all object pixel features rather than pooling them into a single vector because our goal is to densely predict depths for the entire region of objects. 
On the basis of parsed objects $\mathcal{O}$, we devise a Structural Priors Mining approach (SPM) and a Self-Boosting Learning (SBL) strategy to unleash the potential of objects' inherent properties in the depth estimation from category-centric and instance-centric perspectives, respectively, which will be elaborated in the following parts. 


\subsection{Structural Priors Mining}
\subsubsection{Category-Specific Depth Decoders.}
Estimating depths from monocular images is challenging as it requires understanding the relationships between physical scales and depth values of objects with different semantics. Existing BEV-based methods~\cite{li2023bevdepth,li2022bevstereo} utilize prevalent pre-trained image backbones~\cite{he2016deep,liu2022convnet} as feature encoders to endow the model with strong semantic-capturing ability, however, they rely on a single depth decoder to simultaneously learn scale-to-depth mapping patterns of multiple semantic categories, increasing the burden of optimization. 

To simplify learning such patterns of different semantic categories, we design multiple parallel lightweight depth decoders, where each of them is responsible for processing objects of the same category as shown in Fig~\ref{fig2}. Specifically, we first divide all foreground objects into several non-overlapping semantic groups $\{\mathcal{O}^{(c_i)}\}_{i=1}^{K}$, where $K$ is the number of object categories.
Then, taking an object $o^{(c_i)}_{j}=(F^{(c_i)}_{j}, b^{(c_i)}_{j})$ from semantic group $\mathcal{O}^{(c_i)}$ as an example, we feed both object features and box parameters (i.e., normalized box height and width) into a lightweight depth decoder. Within each depth decoder, box parameters are encoded by a linear mapping and then fused with object features using the SE block~\cite{hu2018squeeze}. And the outputs will be fed into a convolutional layer to predict the depth logits $D^{(c_i)}_{j}$ for regions of the current object. The above process can be formulated as: 
\begin{equation}
    \begin{split}
        & \widetilde{F}^{(c_i)}_{j} = \mathtt{SE}(F^{(c_i)}_{j};\mathtt{Linear}(b^{(c_i)}_{j})) \\
        & D^{(c_i)}_{j} = \mathtt{Conv}(\widetilde{F}^{(c_i)}_{j}) \\
    \end{split}
\end{equation}
Finally, by merging the predicted depth logits of all instances, the estimated monocular depth can be obtained.

\subsubsection{Instance-Aware Supervision.}
In a typical BEV-based perception pipeline~\cite{li2023bevdepth}, the depth prediction is supervised by pixel-wise cross-entropy loss, which fails to capture fine-grained instance-level cues, making it more challenging for the aforementioned category-specific depth decoders to learn semantic structural priors. Therefore, we design two new loss functions to encourage learning the rough instance-level absolute depths and inner-instance relative depths. 
First, for object $o^{(c_i)}_{j}$ we convert the discrete depth prediction $D^{(c_i)}_{j} \in \mathbb{R}^{B \times M}$ into continuous depth values $\tilde{D}^{(c_i)}_{j} \in \mathbb{R}^{M}$ following MonoDETR~\cite{zhang2022monodetr}:
\begin{equation}
    \hat{D}^{(c_i)}_{j} = \sum_{k=1}^{B}(d\left[k\right]\cdot \mathtt{Softmax}(D^{(c_i)}_{j}, \mathtt{dim=0})[k])
\end{equation}
where $d\left[k\right]$ represents the depth value of the center of $k$-th depth bin, $B$ and $M$ represent the number of pre-defined depth bins and object pixels. Then, we project the LiDAR points within ground-truth 3D boxes onto the image plane to obtain ground-truth depth values, and keep those intersecting with foreground objects $\mathcal{O}$ to further construct supervision signals. Here, we denote objects with both predicted and ground-truth depths as $\mathcal{O}'=\{(F_{i},b_{i},\hat{D}_i,D^{gt}_i)\}_{i=1}^{N'}$, where $D^{gt}_i \in \mathbb{R}^{M'_{i}}$, $M'_{i}$ is the number of ground-truth depth values. To explicitly supervise the instance-level depth prediction, for each object $o'_{i}=(F_{i},b_{i},\hat{D}_i,D^{gt}_i)$ in $\mathcal{O}'$, we abstract an absolute depth value $d^{gt}_{i}$ from $D^{gt}_{i}$ as the regression target.
It is worth noting that there exist some outliers in the ground-truth depths $D^{gt}_{i}$ due to imperfect sensor calibration~\cite{zhao2023lif}\footnote{Intuitive illustrations of this phenomenon are included in the supplementary materials.}, which poses great challenges in choosing a proper $d^{gt}_{i}$. Therefore, we first scatter all depth values in $D^{gt}_{i}$ into predefined depth bins, and then only average those in the depth bin with the maximum votes as $d^{gt}_{i}$. Afterward, the total absolute depth loss can be calculated as:
\begin{equation}
    \mathcal{L}^{abs}_{depth} = \frac{1}{N'}\sum_{i=1}^{N'}\frac{1}{M'_{i}}\sum_{j=1}^{M'_{i}}(d^{gt}_{i}-\hat{D}_{i}\left[j\right])^{2}
\end{equation}
On the basis of $d^{gt}_{i}$, we also calculate the relative depth loss to encourage the category-specific decoder to learn fine-grained object geometric patterns:
\begin{equation}
    \mathcal{L}^{rel}_{depth} = \frac{1}{N'}\sum_{i=1}^{N'}\frac{1}{M'_{i}}\sum_{j=1}^{M'_{i}}((d^{gt}_{i}-\hat{D}_{i}\left[j\right])-(d^{gt}_{i}-D^{gt}_{i}\left[j\right]))^{2}
\end{equation}

\subsection{Self-Boosting Learning}
The temporal stereo-matching technique relies on geometric consistency through time for depth estimation~\cite{wang2022sts}. Concretely, for every pixel in the $T$-th frame, several depth hypotheses are initially proposed along the depth channel. Then, these hypotheses are warped to the $(T-1)$-th frame to construct cost volume for learning the best match among them. 
In the above process, the main barrier lies in the large memory cost brought by constructing 3D cost volume for huge amounts of pixels in high-resolution image features and dense hypotheses~\cite{li2022bevstereo}. However, the image regions should not be treated equally in our scenario. First, the foreground objects are more important than the background area. Furthermore, it is harder to accurately estimate depth for objects with lower visual clarity and more attention should be paid to them. Therefore, we devise a self-boosting strategy to iteratively focus on harder object regions, which further enables adaptively adjusting the granularity of cost volume construction for different regions and results in a better trade-off between cost and efficacy. 

\subsubsection{Sparse Cost Volume Construction.}
In pursuit of enhanced efficiency, we mainly focus on exploring the stereo-matching behaviors of foreground objects at $T$-th frame, which breaks the conventional dense cost volume construction paradigm. Therefore, we reformulate such a process into a sparse format introduced as follows. Taking a pixel with coordinates $(u,v)$ and depth hypothesis $d^{h}$ as example, we employ the homography warping between $T$-th and $(T-1)$-th frames on it to obtain the corresponding projected location $(u^{T-1},v^{T-1})$: 
\begin{equation}
    (u^{T-1},v^{T-1}) = \mathtt{Homo}((u,v,d^{h});\mathcal{K};\mathcal{M}_{T\rightarrow T-1})
\end{equation}
where $\mathcal{K}$ is the camera intrinsic parameters, and $\mathcal{M}_{T\rightarrow T-1}$ is the transformation matrix from $T$-th to $(T-1)$-th frame. Following the above process, for every object pixel with different depth hypotheses, we establish its correspondence to pixels in $(T-1)$-th frame, and then combine their features to generate the sparse cost volume $V\in \mathbb{R}^{N_p \times N_d \times C_f}$, where $N_p$ and $N_d$ are number of foreground pixels and depth hypotheses, respectively, and $C_f$ is the feature channel dimension. Subsequently, the matching scores are calculated with 3D sparse convolutions~\cite{spconv2022}. 

\begin{table*}[!t]
\centering
\scalebox{0.9}{
\begin{tabular}{l|c|ccccccc}
\toprule
Method            & Input Size & mAP $\uparrow$           & mATE $\downarrow$          & mASE $\downarrow$          & mAOE $\downarrow$          & mAVE $\downarrow$          & mAAE $\downarrow$          & NDS $\uparrow$           \\ \midrule
BEVDet4D-R50~\cite{huang2021bevdet}      & 256x704    & 0.322          & 0.703          & 0.278          & 0.495          & 0.354          & 0.206          & 0.457          \\
PETR-R50~\cite{liu2022petr}          & 384x1056   & 0.313          & 0.768          & 0.278          & 0.495          & 0.923          & 0.225          & 0.381          \\
BEVDepth-R50~\cite{li2023bevdepth}      & 256x704    & 0.351          & 0.639          & 0.267          & 0.479          & 0.428          & 0.198          & 0.475          \\
BEVStereo-R50~\cite{li2022bevstereo}     & 256x704    & 0.372          & 0.598          & \textbf{0.270} & \textbf{0.438} & 0.367          & \textbf{0.190} & 0.500          \\
AeDet-R50~\cite{feng2023aedet}         & 256x704    & 0.387          & 0.598          & 0.276          & 0.461          & 0.392          & 0.196          & 0.501          \\
FB-BEV-R50~\cite{li2023fb}         & 256x704    & 0.378          & 0.620          & 0.273          & 0.444          & 0.374          & 0.200          & 0.498          \\
\textbf{IA-BEV-R50 (ours)} & 256x704    & \textbf{0.400} & \textbf{0.557} & 0.275          & 0.449          & \textbf{0.347} & 0.209          & \textbf{0.516} \\ \midrule
BEVDepth-ConvNeXt-B~\cite{li2023bevdepth}      & 512x1408    & 0.462          & 0.540          & 0.254          & 0.355          & 0.379          & 0.200          & 0.558          \\
BEVStereo-ConvNeXt-B~\cite{li2022bevstereo}      & 512x1408    & 0.478          & -          & -          & -          & -          & -          & 0.575          \\
SA-BEV-ConvNeXt-B~\cite{zhang2023sa}      & 512x1408    & 0.479          & -          & -          & -          & -          & -          & 0.579          \\
\textbf{IA-BEV-ConvNeXt-B (ours)} & 512x1408    & \textbf{0.493}          & \textbf{0.493}          & 0.259          & 0.364          & \textbf{0.336}          & 0.207          & \textbf{0.581}          \\ \bottomrule
\end{tabular}}
\caption{Comparison with state-of-the-art methods on nuScenes val set.}
\label{tbl:SOTA_val}
\end{table*}

\begin{table*}[!h]
\centering
\scalebox{0.9}{
\begin{tabular}{l|c|ccccccc}
\toprule
Method            & Input Size & mAP $\uparrow$           & mATE $\downarrow$          & mASE $\downarrow$          & mAOE $\downarrow$          & mAVE $\downarrow$          & mAAE $\downarrow$          & NDS $\uparrow$           \\ \midrule
BEVDet4D-Swin-B~\cite{huang2021bevdet} & 640x1600    & 0.451          & 0.511          & 0.241          & 0.386          & 0.301          & 0.121          & 0.569          \\
BEVFormer-Vov99~\cite{li2022bevformer} & 640x1600    & 0.481          & 0.582          & 0.256          & 0.375          & 0.378          & 0.126          & 0.569          \\
PETRv2~\cite{liu2022petrv2}  & 640x1600    & 0.490          & 0.561          & 0.243          & 0.361          & 0.343          & 0.120          & 0.582          \\
BEVDepth-ConvNeXt-B~\cite{li2023bevdepth} & 640x1600    & 0.520          & 0.445          & 0.243          & 0.352          & 0.347          & 0.127          & 0.609          \\
BEVStereo-Vov99~\cite{li2022bevstereo} & 640x1600    & 0.525          & 0.431          & 0.246          & 0.358          & 0.357          & 0.138          & 0.610          \\
OA-BEV-Vov99~\cite{chu2023oa} & 900x1600    & 0.494          & 0.574          & 0.256          & 0.377          & 0.385          & 0.132          & 0.575          \\
MV2D-Vov99~\cite{chu2023oa} & 640x1600    & 0.511          & 0.525          & 0.243          & 0.357          & 0.357          & 0.120          & 0.596          \\
SOLOFusion-ConvNeXt-B$^{\dagger}$~\cite{park2022time} & 640x1600    & 0.540          & 0.453          & 0.257          & 0.376          & 0.276          & 0.148          & 0.619          \\
AeDet-ConvNeXt-B~\cite{feng2023aedet} & 640x1600    & 0.531          & 0.439          & 0.247          & 0.344          & 0.292          & 0.130          & 0.620          \\
CAPE-Vov99~\cite{xiong2023cape} & 900x1600    & 0.525          & 0.503          & 0.242          & 0.361          & 0.306          & \textbf{0.114}          & 0.610          \\
FB-BEV-Vov99~\cite{li2023fb} & 640x1600    & 0.537          & 0.439          & 0.250          & 0.358          & \textbf{0.270}          & 0.128          & 0.624          \\
SA-BEV-Vov99~\cite{zhang2023sa} & 640x1600    & 0.533          & 0.430          & \textbf{0.241}          & \textbf{0.338}          & 0.282          & 0.139          & 0.624          \\
\textbf{IA-BEV-ConvNeXt-B (ours)} & 640x1600    & \textbf{0.545}          & \textbf{0.407}          & 0.248          & 0.343          & 0.294          & 0.133          & \textbf{0.630}          \\ \bottomrule
\end{tabular}}
\caption{Comparison with state-of-the-art methods on nuScenes test set. $\dagger$ denotes that longer temporal frames ($>$2) are used.}
\label{tbl:SOTA_test}
\vspace{-8pt}
\end{table*}

\begin{table}[!h]
\centering
\scalebox{0.7}{
\begin{tabular}{c|ccccc}
\toprule
Method        & SILog $\downarrow$ & AbsRel $\downarrow$ & SqRel $\downarrow$ &log10 $\downarrow$ & RMSE $\downarrow$\\ \midrule
BEVDepth~\cite{li2023bevdepth}      & 21.74     & 0.155  & 1.223    & 0.060     & 5.269    \\
BEVStereo~\cite{li2022bevstereo}     & 21.74     & 0.152   & 1.206   & 0.059     & 5.246    \\
IA-BEV (ours) & \textbf{17.64}     & \textbf{0.134}      & \textbf{1.178} & \textbf{0.046}   & \textbf{3.843}    \\ \bottomrule
\end{tabular}}
\caption{Evaluation of depth prediction performances of different methods.}
\label{tbl:SOTA_depth}
\vspace{-8pt}
\end{table}

\subsubsection{Iterative Stereo Matching.}
In the first round, to efficiently recognize objects with rich visual details, we uniformly sample sparse depth hypotheses $H_0 \in \mathbb{R}^{L_0}$ for all pixels $\mathcal{P}_{0}$ in foreground objects $\mathcal{O}$. 
Then the sparse cost volume is constructed based on $\mathcal{P}_{0}$ and $H_0$ to calculate the matching scores $S_0 \in \mathbb{R}^{N_0\times L_0}$, where $N_0$ and $L_0$ are number of pixels and depth hypotheses, respectively. For pixel $\mathcal{P}_{0,i}$, we calculate the mean $\mu_{0,i}$ and standard deviation $\sigma_{0,i}$ along its depth channel as:
\begin{align}
    &\mu_{0,i} = \sum_{j=1}^{L_0}(H_{0}\left[j\right]\cdot S_{0,i}\left[j\right]) \\
    &\sigma_{0,i}^{2} = \sum_{j=1}^{L_0}((H_0\left[j\right]-\mu_{0,i})^{2}\cdot S_{0,i}\left[j\right])
\end{align}
The scale of $\sigma_{0,i}$ indicates the uncertainty of the stereo depth estimation. With small $\sigma_{0,i}$, the depth hypotheses have been successfully verified to find the best match. Conversely, large $\sigma_{0,i}$ means that multiple depth hypotheses are preferred, and thus should be further boosted. Therefore, we regard the pixels whose matching scores'  standard deviation are smaller than our predefined threshold $\sigma_t$ as satisfactory results, and filter them in the next iteration. For the remaining pixels, their mean and standard deviation can provide a more accurate search range, which facilitates proposing depth hypotheses more effectively for the next iteration. With $\mu_{0}$ and $\sigma_{0}$, we update the depth sampling range as:
\begin{equation}
    \mathcal{R}_{1}=\left[\mu_{0}-3\sigma_{0}, \mu_{0}+3\sigma_{0} \right]
\end{equation}
Within $\mathcal{R}_{1}$, we further uniformly sample $L_1$ ($L_1 > L_0$) depth hypotheses $H_1$ for remaining pixels $\mathcal{P}_{1}$. Both $H_1$ and $\mathcal{P}_{1}$ will be utilized for constructing the sparse cost volume and calculating mean and deviation similarly in the next iteration. Since the numbers of depth hypotheses in different iterations are different, we employ an interpolation operation to fill all predefined depth bins for alignment. As shown in Fig.~\ref{fig_case}, the proposed self-boosting learning strategy can distinguish the main regions of clear objects in the early iteration, to save resources to emphasize unclear objects.

\subsection{BEV Feature Processor}
By summing up the predicted monocular and stereo depths from SPM and SBL, the final depth prediction can be obtained for rendering the BEV feature from multi-camera images. Afterward, the BEV feature will be fed into a conventional detection head for the ultimate 3D detection. The total loss functions can be formulated as:
\begin{equation}
    \mathcal{L} = \lambda_{1}\mathcal{L}_{det} + \lambda_{2}\mathcal{L}^{CE}_{depth} + \lambda_{3}\mathcal{L}^{abs}_{depth} + \lambda_{4}\mathcal{L}^{rel}_{depth}
\end{equation}
where $\mathcal{L}_{det}$ and $\mathcal{L}^{CE}_{depth}$ are conventional detection loss and pixel-wise cross-entropy depth loss, respectively.

\section{Experiments}
\subsection{Experimental Setup}
\subsubsection{Dataset and Metrics}
NuScenes dataset~\cite{caesar2020nuscenes} is a large-scale autonomous driving benchmark, encompassing LiDAR, camera and radar data collected from 10,000 unique driving scenarios. These scenarios are systematically grouped into 700 for training, 150 for validation, and 150 for testing. For the purpose of detection, a suite of evaluation metrics are introduced, including the nuScenes Detection Score (NDS), mean Average Precision (mAP), alongside five True Positive (TP) metrics, specifically, mean Average Translation Error (mATE), mean Average Scale Error (mASE), mean Average Orientation Error (mAOE), mean Average Velocity Error (mAVE), and mean Average Attribute Error (mAAE). For depth estimation quality evaluation, the scale-invariant logarithmic error (SILog), mean absolute relative error (AbsRel), mean squared relative error (SqRel), mean log10 error (log10) and root mean squared error (RMSE) are reported.

\subsubsection{Implementation Details}
We use the BEVDepth~\cite{li2023bevdepth} as the baseline method, and follow its official training configurations, including data augmentation, optimizer selection, and basic hyperparameters. Within SPM, objects are divided into 6 semantic groups: \textit{``car", ``truck \& construction vehicle", ``bus \& trailer", ``barrier", ``motorcycle \& bicycle", ``pedestrian \& traffic cone"}. As for SBL, we perform the sparse stereo matching for 2 rounds. The number of sampled depth hypotheses in the first iteration ($L_0$) and second iteration ($L_1$) are 12 and 20, respectively.
In the loss function, the balance factors $\lambda_1$, $\lambda_2$, $\lambda_3$ and $\lambda_4$ are set as 1.0, 3.0, 0.5, 2.0, respectively.
We equip our IA-BEV with both ResNet-50~\cite{he2016deep} and ConvNeXt-base~\cite{liu2022convnet} as image backbones. Input resolutions are rescaled to 256$\times$704, 512$\times$1408 and 640$\times$1600 for comprehensive evaluation.

\subsection{Comparison with State-of-the-art Methods}
In this section, we train our IA-BEV with different configurations for 20 epochs using both CBGS~\cite{zhu2019class} and EMA techniques following prior works~\cite{li2023bevdepth,feng2023aedet,zhang2023sa}.
First, we compare the 3D object detection performances of our IA-BEV with state-of-the-art methods on both nuScenes \emph{val} set and \emph{test} set with different input resolutions and image backbones in Table~\ref{tbl:SOTA_val} and Table~\ref{tbl:SOTA_test}. It can be observed that our IA-BEV consistently surpasses state-of-the-art methods across different configurations on both \emph{val} and \emph{test} sets. Furthermore, compared with OA-BEV~\cite{chu2023oa} and MV2D~\cite{wang2023object} methods which also leverage 2D instance priors, our IA-BEV outperforms them by a clear margin.
Besides, we also compare the depth estimation quality in Table~\ref{tbl:SOTA_depth}. With the help of instance awareness, our IA-BEV evidently enhances the depth estimation quality, which is key to the effectiveness of our method.

\begin{table}[!t]
\centering
\scalebox{0.9}{
\begin{tabular}{cc|cccc}
\toprule
SPM & SBL & mAP $\uparrow$           & mATE $\downarrow$          & mAVE $\downarrow$          & NDS $\uparrow$           \\ \midrule
    &     & 0.330          & 0.700          & 0.552          & 0.425          \\
\checkmark   &     & 0.345          & 0.671          & 0.521          & 0.443          \\
    & \checkmark   & 0.354          & 0.667          & 0.512          & 0.446          \\
\checkmark   & \checkmark   & \textbf{0.367} & \textbf{0.658} & \textbf{0.486} & \textbf{0.461} \\ \bottomrule
\end{tabular}}
\caption{Ablation study of the proposed SPM and SBL.}
\label{tbl:abl_main}
\end{table}

\begin{table}[!h]
\centering
\scalebox{0.9}{
\begin{tabular}{c|cccc|ccc}
\toprule
\# &PD & CD & $\mathcal{L}^{abs}_{depth}$ & $\mathcal{L}^{rel}_{depth}$ & mAP $\uparrow$ & mATE $\downarrow$  & NDS $\uparrow$  \\ \midrule
1   &    &    &    &  &0.330 & 0.700 & 0.425 \\
2 &\checkmark  &    &    &    & 0.329 & 0.697 & 0.426 \\
3 &  & \checkmark  &    &    & 0.336 & 0.689 & 0.431 \\
4 &  & \checkmark  & \checkmark  &    & 0.340 & 0.691 & 0.439 \\
5 &  & \checkmark  & \checkmark  & \checkmark  & \textbf{0.345} & \textbf{0.671} & \textbf{0.443} \\ \bottomrule
\end{tabular}}
\caption{Ablation study of designs in SPM. ``PD" and ``CD" are short for ``Parallel Decoders" and ``Category-specific decoders". Parallel decoders have the same model structure as category-specific decoders, but take all objects as inputs for each branch.}
\label{tbl:abl_SPM}
\end{table}

\begin{table}[!h]
\centering
\scalebox{0.9}{
\begin{tabular}{cc|ccc}
\toprule
Iter\_num & Memory & mAP $\uparrow$  & mATE $\downarrow$ & NDS $\uparrow$  \\ \midrule
0         & 4.56G    & 0.330 & 0.700 & 0.425 \\
1         & 4.83G    & 0.346 & 0.676 & 0.439 \\
2         & 4.84G    & 0.354 & 0.667 & 0.446 \\
3         & 4.85G    & \textbf{0.356} & \textbf{0.655} & \textbf{0.447} \\ \bottomrule
\end{tabular}}
\caption{Memory cost and performance comparison of iterating different rounds in SBL. The memory costs are based on our reproduction, which are similar to the memory costs measured in SOLOFusion~\cite{park2022time}.}
\label{tbl:abl_SBL}
\end{table}

\subsection{Comprehensive Analysis}
In this section, all models are trained for 24 epochs without using CBGS or EMA techniques for efficient evaluation. The baseline method is the BEVDepth~\cite{li2023bevdepth} without using its ``Depth Refinement" module.
\subsubsection{Ablations of main components.}
We verify the effects of the proposed Structural Priors Mining (SPM) and Self-Boosting Learning (SBL) approaches as shown in Table~\ref{tbl:abl_main}. By introducing SPM or SBL alone can bring 1.8\% and 2.1\% NDS improvements over the baseline method, respectively. And combining them can further boost NDS from 42.5\% to 46.1\%. The significant performance improvements demonstrate the effectiveness of our designs.

\begin{figure*}[!t]
\centering
\includegraphics[width=0.9\textwidth]{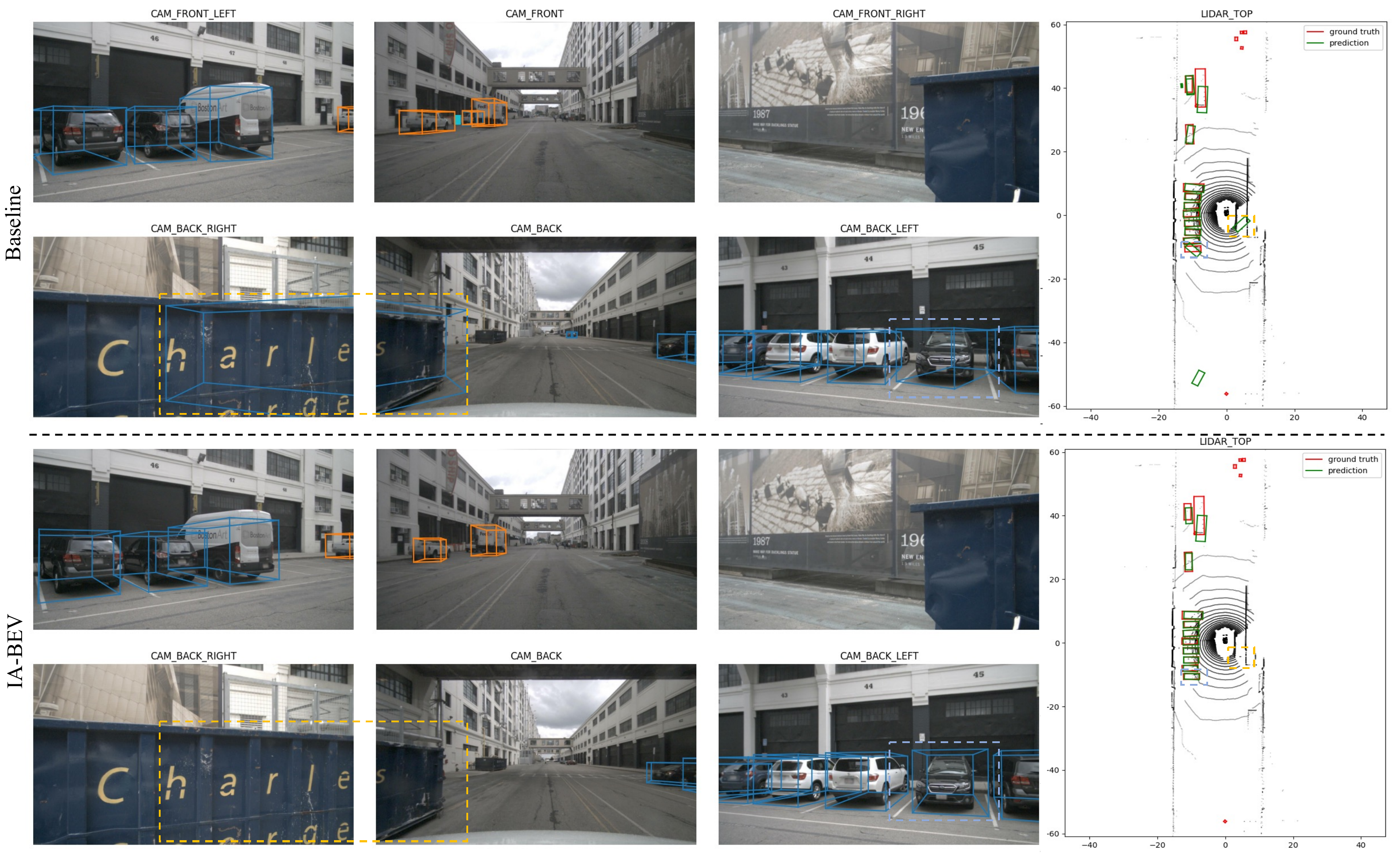} 
\caption{Qualitative results comparison between our IA-BEV and baseline method (i.e., BEVDepth). The dotted rectangles with the same color in image views and BEV planes represent the same regions.}
\label{fig_qual}
\vspace{-8pt}
\end{figure*}

\subsubsection{Ablations of SPM designs.}
We further investigate the effects of each design in SPM. From the Table~\ref{tbl:abl_SPM}, we have the following two observations. First, simply increasing the number of depth decoders without semantic grouping (\#1) can not bring benefits over the baseline (\#2), while our category-specific depth decoder design (\#3) boosts the performance, showcasing the effectiveness of capturing the instances' structural priors per category. Besides, on the basis of category-specific decoders, adding instance-level absolute depth supervision (\#4) can bring 0.4\% mAP and 0.8\% NDS improvements. And by introducing inner-instance relative depth can achieve further enhancement, indicating that instance-level depth supervision can benefit effective monocular depth estimation.

\subsubsection{Effects of SBL iteration rounds.} To evaluate the effects of self-boosting learning (SBL) design, we compare results of different number of iterations in Table~\ref{tbl:abl_SBL}. Iterating 0 rounds stands for the baseline method without using the stereo-matching technique. 
Compared with only iterating 1 round, boosting unclear regions with the second round can significantly improve the performances while only slightly increase the memory, which demonstrates that our self-boosting learning can efficiently promote comprehensive stereo matching. However, increasing the iteration number to 3 does not bring a significant performance boost, which might be because of the limitation of the current resolution of perception and feature quality. Therefore, we only iterate for 2 rounds in practice.

\subsection{Visualization}
\subsubsection{Qualitative results.}
We illustrate the detection results of the BEVDepth baseline and our IA-BEV in Fig~\ref{fig_qual}. Benefitting from 2D instance awareness, our IA-BEV can predict more accurate results than the baseline. First, as shown in the yellow dotted rectangle, our IA-BEV can predict fewer false positives with 2D semantics as priors. Besides, as shown in the blue dotted rectangle, IA-BEV can also perceive the object's detailed structures more accurately than the baseline.

\subsubsection{Case study of SBL.}
\begin{figure}[!t]
\centering
\includegraphics[width=\columnwidth]{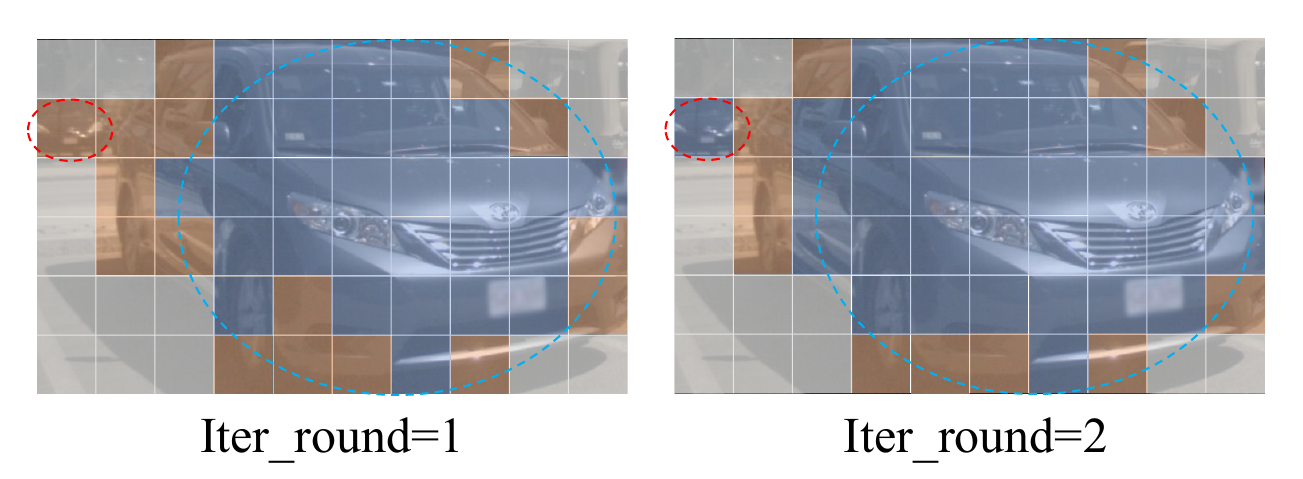} 
\caption{Visualization of filtering and remaining patches in SBL. Gray patches represent background area, and blue and orange patches represent filtered and remaining ones, respectively. We use red and blue dashed circles to highlight vague and clear objects, respectively.}
\label{fig_case}
\vspace{-8pt}
\end{figure}

We also visualize the filtered and remaining patches after the first iteration in SBL in Fig.~\ref{fig_case}. After the first round, the vague car (red circle) is kept by the model for boosting, while the main parts of the closer and clearer car (in the blue circle) are filtered. After the second round of boosting, the vague car has lower stereo-matching uncertainty and is filtered.

\section{Conclusion}
In this paper, we proposed IA-BEV, which enhances the depth estimation process for the multi-camera BEV-based detector by exploring the inherent properties encapsulated in foreground objects. Within IA-BEV, a Structural Priors Mining approach (SPM) and a Self-Boosting Learning strategy (SBL) are proposed to enhance the monocular and stereo depth estimations, respectively. Equipped with both SPM and SBL, IA-BEV sets new state-of-the-art performances across methods that use short-term (i.e., 2) frames with 54.5\% mAP and 63.0\% NDS on nuScenes.

\bibliography{aaai24}

\end{document}